# Automated Detection and Analysis of Minor Deformations in Flat Walls Due to Railway Vibrations Using LiDAR and Machine Learning

Surjo Dey
Computer Science & Engineering
Rajiv Gandhi institute of petroleum technology
Amethi, India
22cs3072@rgipt.ac.in

Ankit Sharma
Computer Science & Engineering
Rajiv Gandhi institute of petroleum technology
Amethi, India
ankits23cs@rgipt.ac.in

Hritu Raj
Computer Science & Engineering
Rajiv Gandhi institute of petroleum technology
Amethi, India
hritur@rgipt.ac.in

Susham Biswas
Computer Science & Engineering
Rajiv Gandhi institute of petroleum technology
Amethi, India
susham@rgipt.ac.in

*Abstract*— This study introduces an advanced methodology for automatically identifying minor deformations in flat walls caused by vibrations from nearby railway tracks. It leverages high-density Terrestrial Laser Scanner (TLS) LiDAR surveys and AI/ML techniques to collect and analyze data. The scan data is processed into a detailed point cloud, which is segmented to distinguish ground points, trees, buildings, and other objects. The analysis focuses on identifying sections along flat walls and estimating their deformations relative to the ground orientation.

Findings from the study, conducted at the RGIPT campus, reveal significant deformations in walls close to the railway corridor, with the highest deformations ranging from 7 to 8 cm and an average of 3 to 4 cm. In contrast, walls further from the corridor show negligible deformations. The developed automated process for feature extraction and deformation monitoring demonstrates potential for structural health monitoring. By integrating LiDAR data with machine learning, the methodology provides an efficient system for identifying and analyzing structural deformations, highlighting the importance of continuous monitoring for ensuring structural integrity and public safety in urban infrastructure. This approach represents a substantial advancement in automated feature extraction and deformation analysis, contributing to more effective management of urban infrastructure.

## I. Introduction

LiDAR (Light Detection and Ranging) technology has transformed urban planning, environmental monitoring, and structural health assessment. By emitting laser pulses to measure distances, LiDAR creates detailed, high-resolution 3D models of surroundings. This research aims to develop an automated method for extracting key features, like buildings, and monitoring their deformations using LiDAR data, with a particular focus on deformations caused by vibrations from railway tracks. Automating these processes is essential for timely and accurate structural health monitoring, thereby enhancing the safety and sustainability of urban infrastructure.

Our methodology depends on the precise alignment of LiDAR data using the Iterative Closest Point (ICP) algorithm. The ICP algorithm is crucial for minimizing the differences between corresponding points in source and target point clouds, ensuring accurate data alignment. This alignment is fundamental for subsequent analyses, enabling reliable feature extraction and deformation detection.

After aligning the data, the next step is to differentiate between ground and non-ground areas within the LiDAR data. This is done by computing a 3D Delaunay triangulation, which defines the outer boundary and internal structure of the points. By examining the internal angles and edge lengths of the triangles formed, ground areas—typically smoother and more continuous—are distinguished from non-ground areas like buildings and vegetation, which appear more irregular or elevated. Visualizing this data through mesh plots provides a clear representation of the identified features, aiding accurate extraction and analysis.

The primary focus of this research is detecting structural deformations, particularly in buildings near railway tracks, which are susceptible to vibrational impacts. The point cloud data is segmented into vertical sections along the Z-axis to analyze potential deformations. Each segment is inspected for changes in area and structure, with significant variations indicating deformation. By dividing the data into slices and applying linear regression to fit lines in the X-Z plane, deviations from expected trends can be detected, signaling structural anomalies or deformations.

Machine learning algorithms further enhance the deformation detection process by analyzing the segmented data to identify subtle changes that signify potential deformations. This approach is especially effective in detecting deformation patterns in buildings affected by vibrations from nearby railway tracks. For instance, by calculating the slope and intercept for each segment, the algorithm quantifies the extent of deformation and pinpoints specific areas of concern.

The analysis revealed that buildings close to railway tracks showed noticeable deformations due to the vibrations from passing trains, while buildings farther away did not exhibit significant deformations, highlighting the localized impact of railway vibrations. This finding underscores the necessity of continuous monitoring to maintain structural integrity and ensure public safety.

In conclusion, this research showcases the effectiveness of integrating LiDAR data with machine learning algorithms to automate feature extraction and deformation monitoring. The focus on vibrational deformations caused by railways provides valuable insights into the specific impacts of such vibrations on nearby buildings. This automated approach not only improves the efficiency of urban infrastructure management but also offers a scalable solution for large-scale monitoring projects. By detailing the methodology and analysis, this study advances automated structural health monitoring techniques, contributing to safer and more resilient urban environments.

## II. DATA COLLECTION

For this study, high-density point cloud data of building near railway tracks and similar building far away from railway tracks was collected using a Terrestrial Laser Scanner (TLS). [1] TLS provides precise 3D measurements, capturing detailed geometries of structures. The initial dataset includes not only the buildings but also extraneous elements like trees, ground surfaces, and lamp posts, which need to be filtered out to focus on the buildings' structural details. [2]

To combine these multiple scans into a single, cohesive point cloud, we employed the Iterative Closest Point (ICP) algorithm. The ICP algorithm iteratively refines the alignment of point clouds by minimizing the sum of squared differences between corresponding points.

By applying the ICP algorithm, we ensured that the point clouds from different scans were accurately aligned, resulting in a unified point cloud dataset that represents the buildings with high fidelity. This unified dataset was then used for further analysis, including feature extraction and deformation monitoring.

### A. Site Preparation and Scanner Setup
We begin with thorough planning. A site visit determines the optimal location and orientation for the TLS scanner to ensure complete coverage of the target building(s). Once positioned, the scanner's internal sensors undergo calibration for accurate data collection.

### B. Strategic Marker Placement
Markers are strategically placed throughout the environment. These markers play a crucial role in data processing. Spherical markers, with their simple shape, help precisely align and combine individual scans into a cohesive whole. Additionally, checkerboard markers provide numerous reference points for even more precise alignment, ensuring the final point cloud accurately reflects the scanned building.

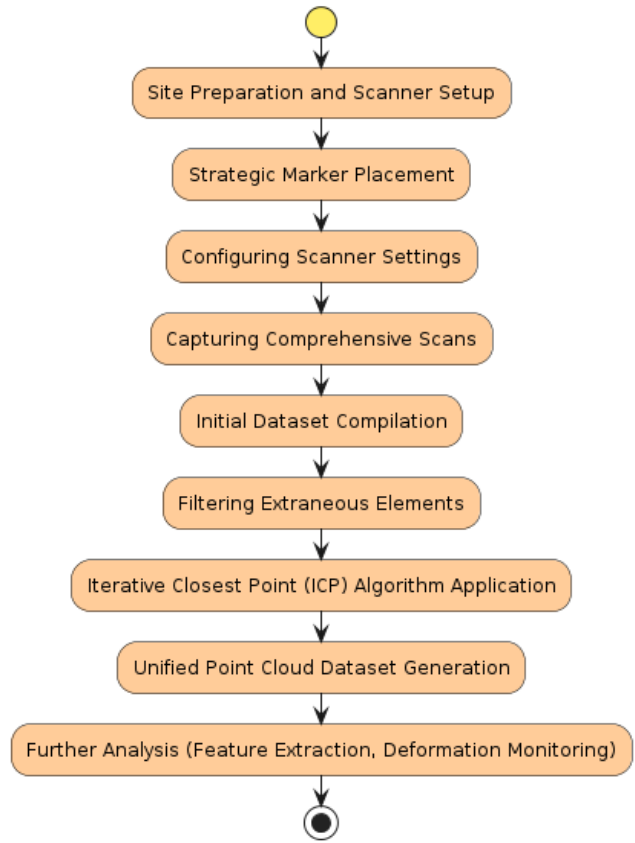

*Figure 1 : Systematic flow for data collection and processing*

### C. Configuring Scanner Settings
With scanner placement and marker deployment finalized, the user configures the scanner settings. These settings, including resolution, scan range, and scan speed, are crucial for optimizing data collection for the specific building being scanned. Resolution determines the level of detail captured, with higher resolutions resulting in finer point clouds but larger data files. Scan range defines the maximum distance the scanner can effectively capture data, while scan speed dictates how quickly the scanner acquires data from a specific point. Selecting the right combination of these parameters ensures efficient and accurate data acquisition tailored to the building's size and complexity.

### D. Capturing Comprehensive Scans
The data collection phase involves capturing multiple scans of the target building. The TLS operator ensures complete coverage by strategically positioning the scanner and capturing scans from various viewpoints. Overlapping sections between scans are essential for facilitating the subsequent merging process.

### III. Scans Processing

After data collection, the individual scans undergo processing to transform them into the required format suitable for further analysis. Here is an overview of our data processing steps:

*1) Stitching Scans*

Specialized software utilizes strategically placed markers to precisely align and merge (register) individual scans into a single coordinate system. This stitching process ensures a seamless and accurate representation of the entire scanned building. The markers act as reference points, allowing the software to piece together the scans with high precision, minimizing errors and overlaps. This step is crucial for creating a coherent and unified model of the building, which forms the foundation for subsequent analysis. [3]

*2) Registration Refinement*

Building upon the initial alignment, the software may employ additional registration techniques to further refine the alignment between scans. This can involve algorithms that analyse the overall geometry and point cloud features to optimize the merging process. Techniques such as Iterative Closest Point (ICP) and other advanced matching algorithms are often used to improve the precision of the alignment. These methods help in reducing any residual errors and ensure that the merged scans accurately reflect the true dimensions and layout of the building.

*3) Generating the Point Cloud*

Once the scans are meticulously aligned and registered, the software processes the merged data to generate a detailed point cloud. This point cloud serves as a digital representation of the entire building's 3D structure, capturing the precise XYZ coordinates of each point. The point cloud data is dense and detailed, providing a comprehensive and accurate 3D model that includes every minute detail of the building's surface. This high-resolution data is essential for conducting thorough and accurate structural analyses. [4]

*4) Exporting XYZRGB Data*

Finally, the software allows for exporting the generated point cloud data in the XYZRGB file format. This format includes not only the spatial coordinates (XYZ) but also the color information (RGB) associated with each point within the point cloud. The inclusion of color information enhances the visual representation of the building, making it easier to identify different materials and features. This rich data allows for further analysis and visualization of the captured 3D environment of the building, facilitating detailed inspections, simulations, and presentations. The XYZRGB data format is widely compatible with various software tools, ensuring that the data can be easily utilized for different applications and purposes. [5]

### IV. Methodology

A. *Data Preprocessing*
- The raw point cloud data undergoes pre-processing to enhance its quality for deformation analysis. This might involve reducing the number of data points (down sampling) for efficient processing and removing outliers that could skew the analysis.

B. *Ground Separation and Non-Ground Classification*
- A crucial step in pre-processing is segmentation, where we separate ground points from non-ground points. Ground points typically represent the underlying terrain or surface, while non-ground points encompass the objects relevant to our study: buildings, trees, and lamp posts.

C. *Isolating Buildings using KNN*
- Our primary focus lies in analyzing building structures for potential deformations. To achieve this, we leverage a K-Nearest Neighbors (KNN) classification algorithm on the non-ground points. The KNN classifier analyzes the features of each non-ground point, such as height, intensity (brightness), and local curvature. By comparing these features to the features of its k nearest neighbors in the data set (using k=17), the KNN classifier assigns a class label to the point. In our case, the class labels would be "building," "tree," or "lamp post." By using KNN we can perfectly confirm the building. [6]

D. *Removing Outliers and Refining Data*
- The classification process might generate some outliers – points that deviate significantly from their assigned class. These outliers could arise due to factors like sensor noise or occlusions in the environment. To ensure accurate analysis of building structures, we meticulously identify and remove these outliers from the classified point cloud data. This refinement step helps us isolate high-confidence building points for further analysis. [7]

E. *Building Separation using RGB Information*
- Having separated ground points and classified non-ground points into building, tree, and lamp post categories, we can leverage the RGB information (color data) within the point cloud to further refine the building data. By analyzing the RGB values of each classified point, we can distinguish building materials (concrete, brick, etc.) from the color signatures of trees and lamp posts.

## V. ANALYZING BUILDING COLUMNS FOR DEFORMATIONS WITH TLS

### A. Input: Building Point Cloud Data

- The process begins with the point cloud dataset representing the building. Then Random Sample Consensus (RANSAC) is applied to the building points which represents building wall.

### B. Selecting a Column for Analysis

- A column from the building point cloud is selected by segmenting the wall into vertical sections or slices for detailed analysis and modeling. This might involve choosing a column near a railway track or one suspected to be more vulnerable to vibrations.

### C. Making Horizontal Slices

- Once a column is selected, the point cloud for that column is virtually sliced horizontally into multiple thin layers, from bottom to top. Imagine creating a layered cake representation of the column from the point cloud data.

#### 1) Compare Slices

Each horizontal slice is then compared to a reference slice, which is typically the bottommost slice. This comparison aims to identify potential deformations within the column. Deformations might manifest as deviations in height, intensity (brightness), or curvature when comparing slices. Significant deviations between corresponding points in different slices could indicate that the column has deformed over time due to vibrations.

#### 2) Greater Than Deformation Threshold?

The comparison process likely involves a threshold of 5 cm for determining significant deviations. Additionally, for vertical deformations in a 25-meter building, the allowable axial shortening threshold due to loading (such as dead loads and live loads) can typically range from about 3 cm to 5 cm. If the difference between corresponding points in two slices exceeds these thresholds, it might be flagged as a potential deformation.
Yes (Deformation) / No (Not Deformation)
Depending on the comparison outcome (greater than the 5 cm horizontal threshold or the vertical threshold for axial shortening, or not), the method would categorize that specific comparison between slices as either indicating a deformation or not.

#### 3) Output: Deformation Analysis Results

After analyzing all relevant building columns, the program outputs the results of the deformation analysis for the chosen column(s). This might include visualizations or data tables highlighting potential deformation zones within the column. By comparing TLS data from potentially affected columns near railways to control columns from buildings further away, we got valuable insights into the impact of train vibrations on specific building elements.

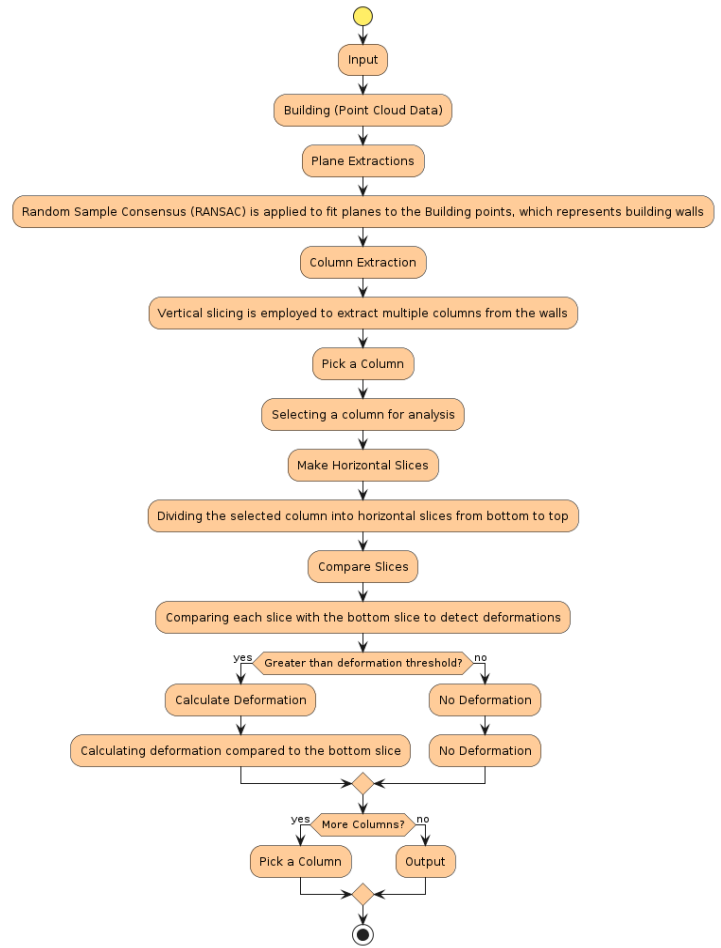

*Figure 2 Flow chart for deformation analysis.*

## VI. RESULT

In the results section, the effectiveness of the developed methodology in automating feature extraction and deformation monitoring using LiDAR data is vividly illustrated through a series of comprehensive visualizations and analyses. Six key images provide valuable insights into different facets of the study. The figure 4 presents the Delaunay triangulation, visually delineating ground and non-ground areas within the LiDAR data, aiding in understanding the spatial distribution of features such as buildings and vegetation. Following this, figure 5 in the form Ground Separation and Non-Ground Classification Subsequently, figures 8, 9 and 10 exhibit the process that taking a sample and slice it in 15 parts. Make a line by using y=mx+c for getting y value and continue doing it in every slice. Figure 8 showcases the squared differences of all 15 columns, highlighting areas of significant deviation that may indicate potential deformations, thereby aiding in pinpointing regions within building structures requiring further investigation. Finally, figure 12 presents a deformation graph, providing a

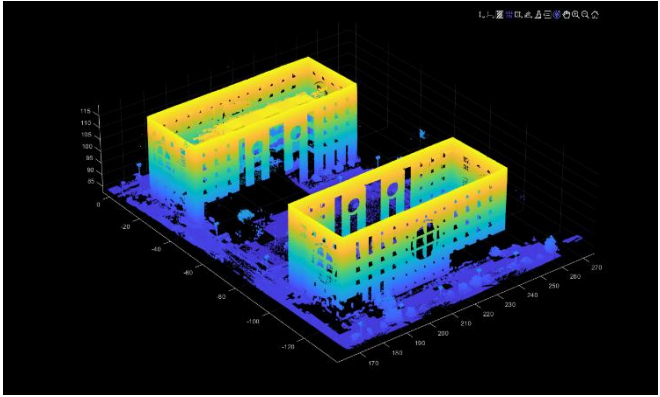

*Figure 3 visualized the point cloud data*

areas within LiDAR data for spatial understanding. comprehensive overview of structural deformations detected during the study and offering insights into their extent and distribution, particularly in buildings near railway tracks. Together, these visualizations and analyses underscore the efficacy of the developed methodology in automating feature extraction and deformation monitoring using LiDAR data, providing valuable insights for informed decision-making in urban infrastructure management and ensuring public safety.

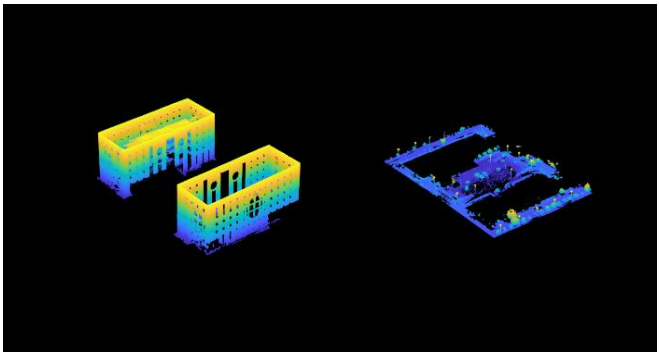

*Figure 4 Ground Separation and Non-Ground Classification*

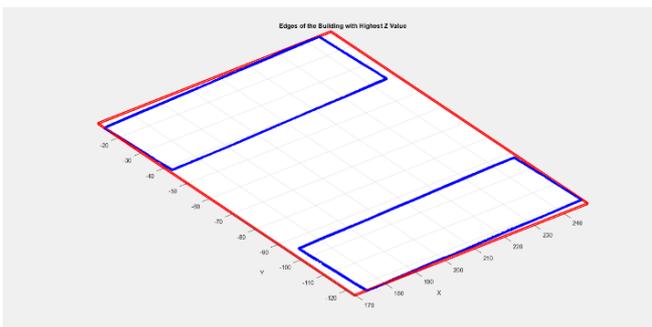

*Figure 5 Isolating Buildings Using KNN*

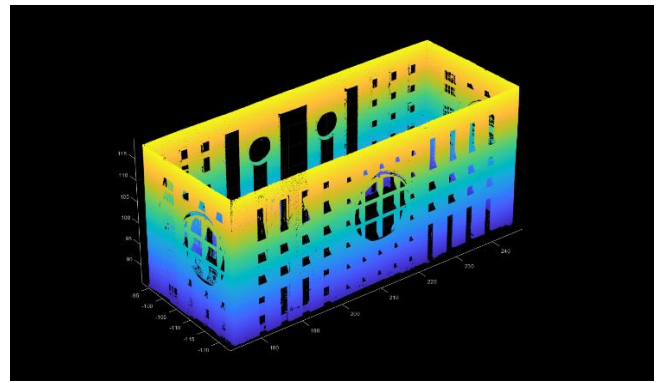

*Figure 6 Removing Outliers and Refining Data*

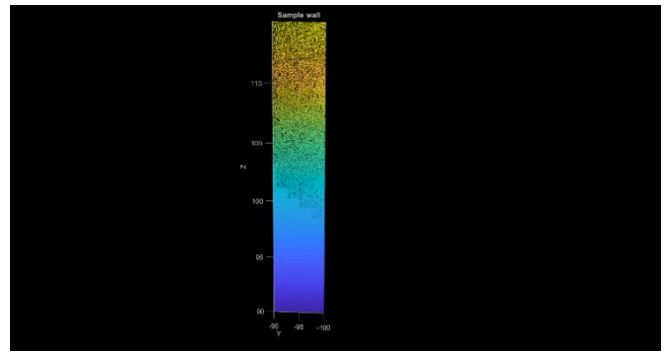

*Figure 7 Taking a sample from the wall*

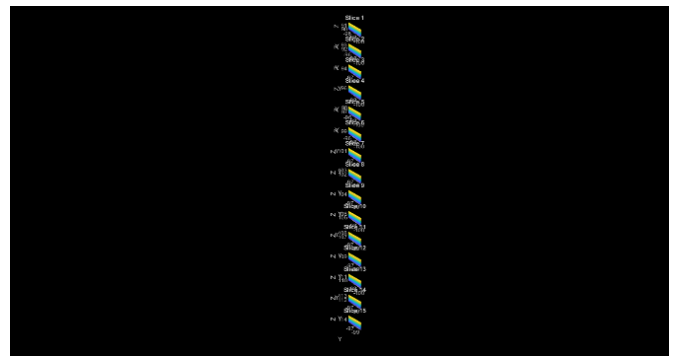

*Figure 8 Slice sample in 15 parts*

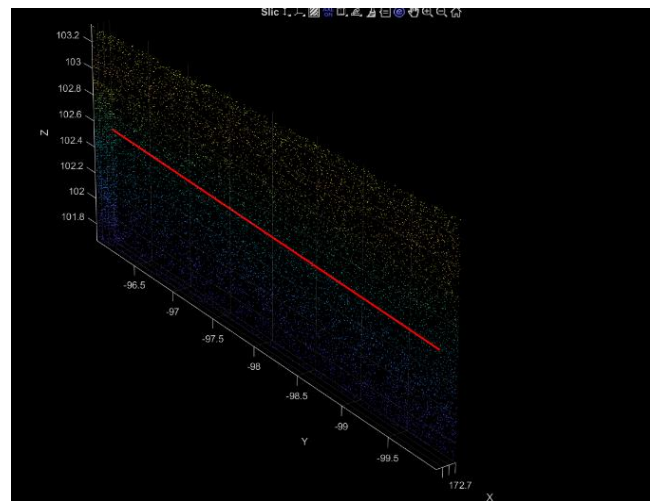

*Figure 9 Make a line by using y=mx+c for getting y value and continue doing it in every slice.*

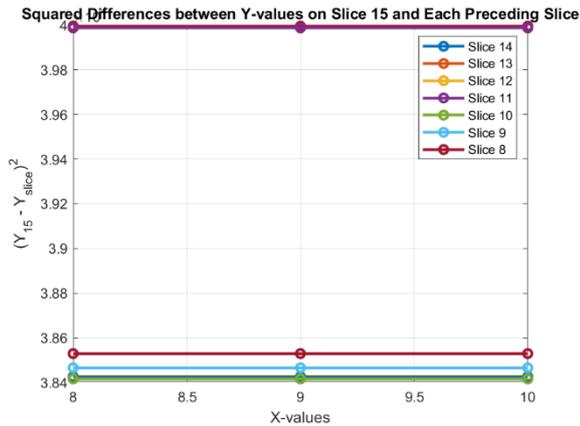

*Figure 10 The Y line all come to reference ground one because of same x value*

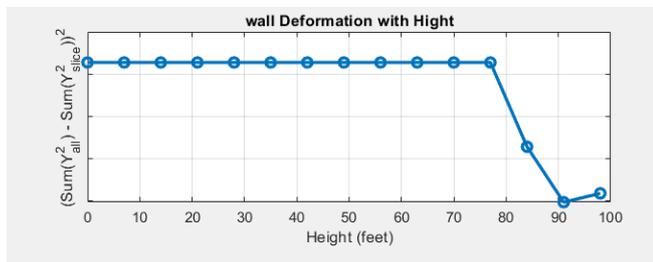

*Figure 11 Deformation Graph*

## VII. CONCLUSION

This analysis provides a valuable tool for assessing the potential effects of train vibrations on building components. By systematically analyzing TLS-derived point cloud data, we can identify potential deformations and make informed decisions regarding building maintenance or mitigation strategies.

**Comparative Analysis of Deformation Detection Methods**

Existing research has utilized LiDAR technology to analyze deformations in visibly cracked and old walls, focusing on structures with clear, observable deformations. Other studies have examined long-building deformations using aerial LiDAR scans, providing a broad overview of structural shifts. While these methods effectively identify significant and apparent deformations, they often overlook subtler structural changes that are not visible to the naked eye or detectable through aerial scans. This limitation can result in undetected early-stage damage, which is crucial for preventive maintenance. [8] [9] [10]

In contrast, our methodology leverages high-density LiDAR data and advanced machine learning algorithms to detect deformations not visible to the naked eye. By segmenting the point cloud data into fine vertical slices and applying linear regression, our process can identify even the slightest deviations from the expected structural form. This detailed approach ensures a comprehensive analysis of the building's integrity, capturing minor deformations that could indicate early signs of structural stress. The automated nature of our method also enhances efficiency and repeatability, making it a more robust solution for continuous structural health monitoring compared to traditional visual or aerial LiDAR methods.